\documentclass{article}

\usepackage{microtype}
\usepackage{graphicx}
\usepackage{subfigure}
\usepackage{booktabs} 
\usepackage{hyperref}

\usepackage[accepted]{icml2023}

\usepackage{amsmath}
\usepackage{amssymb}
\usepackage{mathtools}
\usepackage{amsthm}
\usepackage{setspace}

\usepackage[capitalize,noabbrev]{cleveref}

\theoremstyle{plain}

\theoremstyle{definition}

\theoremstyle{remark}

\usepackage[disable,textsize=tiny]{todonotes}
\usepackage[textsize=tiny]{todonotes}
\newcommand{\chulin}[1]{#1}

\renewcommand{\paragraph}[1]{\textbf{#1}\ \ }

\newcommand{\algname}{FedSelect}

\icmltitlerunning{{\algname}: Customized Selection of Parameters for Fine-Tuning during Personalized Federated Learning}

\begin{document}

\twocolumn[
\icmltitle{FedSelect: Customized Selection of Parameters for Fine-Tuning during Personalized Federated Learning}



\icmlsetsymbol{equal}{*}

\begin{icmlauthorlist}
\icmlauthor{Rishub Tamirisa}{AI_UIUC,UIUC}
\icmlauthor{John Won}{AI_UIUC,UIUC,POSTECH}
\icmlauthor{Chengjun Lu}{AI_UIUC,UIUC}
\icmlauthor{Ron Arel}{AI_UIUC,UIUC}
\icmlauthor{Andy Zhou}{AI_UIUC,UIUC}
\end{icmlauthorlist}

\icmlaffiliation{UIUC}{University of Illinois at Urbana-Champaign}
\icmlaffiliation{AI_UIUC}{Lapis Labs}
\icmlaffiliation{POSTECH}{Department of Computer Science and Engineering, POSTECH}

\icmlcorrespondingauthor{Rishub Tamirisa}{rishubt2@illinois.edu}

\icmlkeywords{Machine Learning, ICML}

\vskip 0.3in
]



\printAffiliationsAndNotice{}  

\begin{abstract}
Recent advancements in federated learning (FL) seek to increase client-level performance by fine-tuning client parameters on local data or personalizing architectures for the local task. Existing methods for such personalization either prune a global model or fine-tune a global model on a local client distribution.  However, these existing methods either personalize at the expense of retaining important global knowledge, or predetermine network layers for fine-tuning, resulting in suboptimal storage of global knowledge within client models. Enlightened by the lottery ticket hypothesis, we first introduce a hypothesis for finding optimal client subnetworks to locally fine-tune while leaving the rest of the parameters frozen. We then propose a novel FL framework, FedSelect, using this procedure that directly personalizes \textit{both client subnetwork structure and parameters}, via the simultaneous discovery of optimal parameters for personalization and the rest of parameters for global aggregation \textit{during training}. We show that this method achieves promising results on CIFAR-10.
\end{abstract}

\begin{spacing}{0.95}

\section{Introduction}
\label{submission}

Federated Learning (FL) \citep{mcmahan2017} is a machine learning paradigm which utilizes multiple clients that collaborate to train models under the supervision of a central aggregator, usually referred to as the server. Unlike traditional centralized methods which require the assemblage of data at the central server, FL methods require that only parameter updates are communicated in order to coordinate the FL training process, such that any number of clients can learn from the decentralized data without direct transfer of data. This allows for maintenance of local data privacy while also providing stronger model performances than what participants could have achieved locally. Accordingly, FL has been adapted to many privacy-sensitive tasks, such as medical data classification \citep{sheller2020federated}. One of the main challenges in FL is the presence of data heterogeneity, where clients' local data distributions vary significantly from one another. 

This problem of data heterogeneity is most commonly addressed by \textit{personalized federated learning (pFL)}, which adapts clients models to local distributions. Most techniques use \textit{full model personalization}, where clients train both a personalized and a global model. However, this requires twice the computational cost of standard FL \citep{dinh2020personalized,li2021ditto} and is impractical in some settings. \textit{Partial model personalization} alleviates this by splitting clients into shared and personalized parameters \citep{pillutla2022federated}, where only the shared parameters are updated globally, but typically results clients that overfit to local distributions and reduced performance. Additionally, the personalized architecture needs to be manually designed before training and cannot be adapted to specific settings.

To address this, we propose \textit{{\algname}}, where we adapt both both \textit{architecture and parameters} for each client to its local distribution during training. Our method is based on the intuition that individual client models should choose only a necessary subset of shared parameters to encode global information  for their local task, since it may not be optimal to reuse all global information from any full layer(s).
We achieve this through the \textit{Lottery Ticket Hypothesis (LTH)}, originally proposed to prune models by finding optimal subnetworks, or lottery ticket networks (LTNs) \citep{frankle2019lottery}. However, instead of pruning the remaining parameters to zero, we reuse them as personalized parameters. We observe improved performance on CIFAR-10 compared to pruning-based LTH-FL approaches \cite{li2020lotteryfl,10.1007/978-3-031-19775-8_5} and other personalized FL approaches \cite{liang2020think,arivazhagan2019federated,collins2021exploiting,li2021ditto,oh2022fedbabu} as well as reduced communication costs compared to partial model personalization.

\paragraph{Related Works.}
Partial model personalization seeks to improve the performance of client models by altering a subset of their structure or weights to better suit their local tasks. It also addresses the issue of ``catastrophic forgetting'' \citep{McCloskey1989CatastrophicII}, an issue in personalized FL where global information is lost when fine-tuning a client model on its local distribution from a global initialization \citep{Kirkpatrick_2017, pillutla2022federated}. It does this by forcefully preserving a subset of parameters, $u$, to serve as a fixed global representation for all clients. However, existing methods introduced for partial model personalization \citep{pillutla2022federated, collins2021exploiting} require hand-selected partitioning of these shared and local parameters, and choose $u$ as only the input or output layers for their experiments. 

LotteryFL \citep{li2020lotteryfl} learns a shared global model via FedAvg \citep{mcmahan2017} and personalizes client models by pruning the global model via the vanilla LTH. Importantly, parameters are pruned to zero according to their magnitude after an iteration of batched stochastic gradient updates. However, due to a low final pruning percentage in LotteryFL, the lottery tickets found for each client share many of the same parameters, and lack sufficient personalization \citep{10.1007/978-3-031-19775-8_5}.

\section{Methods}

\subsection{Problem Definition}

We consider a standard FL setting with $N$ clients and one server. The set $C$ denotes the set of client devices, where $N = |C|$. In particular, $c_k \in C$ denotes the $k$th client whose data distribution is given by $\mathcal{D}_{c_k} = \{x_i^k, y_i^k\}_{i=1}^{N_k}.$ Let $s$ be the number of classes assigned to the clients $c_1,\dots ,c_N$. Next, let $\theta$ denote the vector of parameters defined by the client model architecture. Then the loss of the $k$th client model for each data point $x$ is  $f_k(\theta_k, x)$, where $\theta$ denotes the model parameters. 

While the classical FL objective shown in Equation \ref{eq1} ~\cite{mcmahan2017} seeks to minimize loss across all clients with respect to a global parameter vector $\theta_G$, we focus on partial model personalization.

\begin{equation}
\label{eq1}
    \min_{\theta_G}\frac{1}{N}\sum_{k=1}^N\sum_{i=1}^{N_k}f_k(\theta_G, x_i^k)
\end{equation}

Partial model personalization refers to the procedure in which model parameters are partitioned into shared and local parameters, denoted $u$ and $v$, for averaging and local fine-tuning. 

We consequently define $\theta_k = (u, v_k)$, where $u$ denotes a set of shared global parameters, and $v_k$ the personalized client parameters. The pFL objective following this formulation is given by:
\begin{equation}
    \min_{u, \{v_k\}_{k=1}^N}\sum_{k=1}^N\frac{\alpha_k}{N_k}\sum_{i=1}^{N_k}f_k((u, v_k), x_i^k)
\end{equation}
where $\alpha_k$ represents a constant weighting factor for aggregation of client losses. 




\subsection{Motivation} 

Prior works involving fine-tuning during both transfer learning and federated learning under distributional shift selectively fine-tune models layer-wise~\cite{lee2023surgical, pillutla2022federated, liang2020think,lifedbn, collins2021exploiting}. In this work, we propose a novel hypothesis describing that only \textit{parameters} that change the most during training are necessary for fine-tuning the model; the rest can be frozen as initialized parameters.
Thus, drastic distributional changes in the fine-tuning task may be better accommodated by preserving pretrained knowledge \textit{parameter-wise} rather than \textit{layer-wise}. Following this we propose the following hypothesis:

\textbf{FL Gradient-based Lottery Ticket Hypothesis.} \textit{When training a client model on its local distribution during federated learning, parameters exhibiting minimal variation are considered suitable for freezing and encoding shared knowledge, while parameters demonstrating significant fluctuation are deemed optimal for fine-tuning on local distribution and encoding personalized knowledge.}

The set of parameters selected for personalization will be trained on local data and kept locally, while the rest of the parameters that are identified as frozen will be initialized as the global parameters, then locally updated, and finally submitted to the server for federated averaging to encode shared knowledge across  clients.
GradLTN (Algorithm~\ref{alg:gradltn}) describes the process by which these candidate parameters for local personalization and global updating are identified, respectively. Then, \algname \chulin{(Algorithm~\ref{alg:{\algname}})} utilizes GradLTN's output to perform federated averaging.
\subsection{Algorithms} 
\subsubsection{GradLTN}
\setlength{\textfloatsep}{5mm}
\begin{algorithm}[tb]
   \caption{GradLTN: Gradient-based Lottery Tickets}
   \label{alg:gradltn}
\begin{algorithmic}
   \STATE {\bfseries Input:} $\theta_{0}, L, r$
   \FOR {$i=0$ {\bfseries to} $L$}
        \IF {$i > 0$}
            \STATE $\gamma \gets |\theta_{i} - \theta_{i-1}| \odot m_{i-1}$  \hfill \textit{\textcolor{teal}{\# Find new param change}}
            \STATE $m_i \gets$ binary mask for largest $(1-r)\%$ values in $\gamma$
            $\theta_i \gets \theta_0$ \hfill \textit{\textcolor{teal}{\# Reinitialize model}}
        \ELSE
        \STATE $m_i \gets $ mask of all $1$s
        \ENDIF
        \FOR {epoch $j=1$ {\bfseries to} $E$} 
            \STATE $t \gets 0$
            \FOR {batch $b \in B$}
                \STATE \textit{\textcolor{teal}{\# Freeze params where mask is zero}}
                \STATE $g_t \gets \nabla_{\theta_{i,t}}l(\theta_{i,t}, b) \odot m_i$ 
                \STATE $\theta_{i, t+1} \odot m_i \gets \theta_{i, t} \odot m_i - \eta g_t$ 
            \ENDFOR
        \ENDFOR
   \ENDFOR
   \STATE \textit{\textcolor{teal}{\# Return $u$ and $v$ based on $m_i$}}
   \STATE return $\theta_0 \odot \neg m_L, \theta_0 \odot m_L, m_L$
\end{algorithmic}
\end{algorithm}


GradLTN takes as input an initialization for the network $\theta_0$, the number of total mask-pruning iterations $L$, and a mask-pruning rate $r$. Since statistical heterogeneity is typical across client data distributions in FL, it is a common goal to personalize client architectures or subnetworks to better adapt their local distributions. GradLTN implements this idea during the subnetwork search process, by freezing parameters that change the least, and continually fine-tuning the rest. By the end, two sets of parameters, $\theta_0 \odot \neg m_L$ and $\theta_0 \odot m_L$, are identified for averaging and fine-tuning, respectively. Although we run GradLTN for a fixed number of iterations, there are many alternative choices for the stopping condition, i.e. setting fixed target pruning-rates and target accuracy thresholds.

For convenience, we use the Hadamard operator $\odot$ to be an indexing operator for a binary mask $m$, rather than an elementwise multiply operator. For example, $\theta \odot m$ assumes $\theta$ and $m$ have the same dimensions and returns a reference to the set of parameters in $\theta$ where $m$ is not equal to zero.

\subsubsection{{\algname}}

In {\algname}, the input parameters $C, \theta_G^0, K, R, L, \text{ and } p$ represent clients, the first global initialization, participation rate,  GradLTN iterations, and personalization rate, respectively. The key step in {\algname} is performing LocalAlt on the shared and local parameter partition identified by GradLTN. By the end of GradLTN, $v_k$ is identified as the set of appropriate parameters for dedicated local fine-tuning via LocalAlt; $u$ is also updated in LocalAlt and then averaged for global knowledge acquisition and retention. LocalAlt was introduced to update a defined set of shared and local parameters, $u$ and $v_k$, by alternating full passes of stochastic gradient descent between the two sets of parameters \citep{pillutla2022federated}. To the best of our knowledge, this is the first method to choose parameters for alternating updates in federated learning during training time.

However, averaging among the shared parameters $u_k$ only occurs across parameters for which the corresponding mask entry in $m_k$ is 0. This ensures that only non-LTN parameters are averaged when client models have very different masks. We store a global mask $m_G$ to facilitate updates for clients not sampled during FL. As communication rounds progress, we hypothesize that the global knowledge stored in $\theta_G^t \odot \neg m_i^t$ is refined, and that test accuracy due to personalization will converge.

An important hyperparameter of {\algname} is the personalization rate $p$. For different client problem difficulties \citep{hsieh2020noniid}, the rate of personalization $p$ may affect the level of test accuracy achieved. A nuance of our notation is that for a given $p$, $(1-p) \times 100\%$ parameters are frozen during each of GradLTN's iterations. Therefore we let {\algname} (0.25) denote running {\algname} when 75\% of parameters are frozen in each iteration of GradLTN. So, increasing $p$ corresponds to fewer frozen parameters and greater personalization. Additionally, since only $u_k$ is communicated between the server and clients, a greater $p$ results in reduced communication costs. 

A byproduct of GradLTN is that the subnetwork search process itself fine-tunes parameters in the final iterations of the algorithm, which could be valuable as an initialization for LocalAlt. Therefore, we aim to explore changing the returned values from GradLTN from $\theta_0$ to $\theta_L$ to incorporate this idea.

\setlength{\textfloatsep}{5mm}
\begin{algorithm}[tb]
   \caption{{\algname}}
   \label{alg:{\algname}}
\begin{algorithmic}
   \STATE {\bfseries Input:} $C = \{c_1, \dots, c_N\}, \theta_G^0, K, R, L, p$
   \STATE {\bfseries Server Executes:}
   \begin{ALC@g}
        \STATE $k \gets \max\{{N \cdot K, 1}\}$
        \STATE Initialize all client models $\{\theta_i^0\}_{i=1}^N$ with $\theta_G^0$
        
        \FOR {each round $t$ in $1,2,\dots,R$}
        
            \STATE $S_t \gets$ random sample of $k$ clients from $C$
            \FOR {each client $c_k \in S_t$ {\bfseries in parallel}}
            \STATE \textit{\textcolor{teal}{\# Executed locally on client $c_k$}}
                \STATE $u_k^t, v_k^t, m_k^t \gets$ GradLTN($\theta_k^{t-1}, L, 1-p$) 
                \STATE ${u_k^t}^{+}, {v_k^t}^{+} \gets$ LocalAlt($u_k^t, v_k^t$)
            \ENDFOR
        \STATE \textit{\textcolor{teal}{\# Averaging occurs only across clients where the mask is $1$ for a given parameter's position}}
        \STATE $\theta_G^t \gets$ Average non-LTN parameters $\{{u_k^t}^{+}\}_{c_k}^{S_t}$
        \STATE $m_G^t \gets$ Binary OR over client masks $\bigvee_{c_k}^{S_t}m_k^t$
        \FOR {$i=1$ {\bfseries to} $N$}
            \IF {$m_i^t$ exists}
                \STATE \textit{\textcolor{teal}{\# Distribute global params to clients' non-LTN params, located via $\neg m_i^t$}}
                \STATE $\theta_i^t \odot \neg m_i^t \gets \theta_G^t \odot \neg m_i^t$
                \STATE $\theta_i^t \odot m_i^t \gets {v_i^t}^{+}$
            \ELSE 
                \STATE $\theta_i^t \odot m_G^t \gets \theta_G^t$
            \ENDIF
        \ENDFOR
        \ENDFOR
    \end{ALC@g}
\end{algorithmic}
\end{algorithm}

\vspace{-2mm}
\section{Experiments}

\paragraph{Models \& Datasets.} In this work, we consider a cross-silo setting in which the number of clients is low, but participation is high \citep{liu2022privacy}. We performed all experiments using a ResNet18 \citep{he2015deep} backbone pretrained on ImageNet \citep{5206848}. We show results for our experimental setting on non-iid samples from CIFAR-10. Each client was allocated 20 training samples and 100 testing samples per class. 

\paragraph{Hyperparameters.} We set the number of clients $|C| = N = 10$ in all experiments. The participation rate $K$ is set to 1.0, and the number of classes per client is varied from $s=2$ and $s=4$. In GradLTN, we perform 5 pruning iterations, each with 5 local epochs of training. However, the personalization rate $p$ was varied from 0.25, 0.50, and 0.75. Finally, 5 epochs of personalized training via LocalAlt are performed.
 
\paragraph{Comparisons to Prior Work.} We compare our results to FedAvg \cite{mcmahan2017}, LotteryFL \cite{li2020lotteryfl}, FedBABU \cite{oh2022fedbabu}, FedRep \cite{collins2021exploiting}, FedPer \cite{arivazhagan2019federated}, Ditto \cite{li2021ditto}, and LG-FedAvg \citep{liang2020think}. To fairly compare the performance of these methods, we fix the number of local epochs across all methods to 5. All other hyperparameters (learning rate, momentum, etc.) follow the recommended settings by the authors of the respective works.

\paragraph{Evaluation Metric.} For all methods, the mean accuracy of the final model(s) across individual client data distributions calculated at the final communication round is reported. For FedAvg, accuracy is reported for a single global model. However, for other methods that learn personalized client models, the final average accuracy is reported by averaging individual client model accuracies.

\begin{table}[t]
\vspace{-2mm}
\caption{Mean test accuracies (\%) after 200 communication rounds in the full-participation, low-client setting, with 20 training samples and 100 testing samples per class for each client.}
\label{table1}
\vskip 2mm
\begin{center}
\begin{small}
\begin{tabular}{ lcccr } 
\toprule
Method & $s=2$ & $s=4$\\
\midrule    
FedAvg          & 33.54 & 52.57\\
LG-FedAvg       & 59.95 & 51.35  \\
LotteryFL       & 73.37 & 60.13 \\
Ditto           & 74.50 & 48.85 \\
FedBABU         & 35.65 & 53.03 \\
FedPer          & 30.25 & 51.73 \\
FedRep          & 82.70 & 65.33 \\
FedAlt          & 67.32 & 65.80 \\
{\algname} (0.25)  & 84.22 & 65.23\\
{\algname} (0.50)  & 85.64 & 65.88\\
{\algname} (0.75)  & 85.91 & 62.91 \\
{\algname v2} (e=1)  & 87.46 & 66.51 \\
{\algname v2} (e=3)  & 87.54 & 68.63 \\
{\algname v2} (e=5)  & \textbf{87.99} & \textbf{68.65} \\
\bottomrule
\end{tabular}
\end{small}
\end{center}
\vspace{-2mm}
\end{table}



\subsection{Performance Comparison}
We observe in Table \ref{table1} that {\algname} outperforms all other baselines in the CIFAR-10 in a low-client, full-participation setting. FedAvg learns a global model for all clients, resulting in reduced accuracy when confronted with increased non-IIDness. Conversely, highly personalized FL algorithms exhibit resilience to non-IIDness due to the ease of a client's local task. FedBABU and FedPer appear to suffer the same as FedAvg, despite personalizing a small set of parameters.

\begin{figure}

\begin{center}
\centerline{\includegraphics[width=\columnwidth]{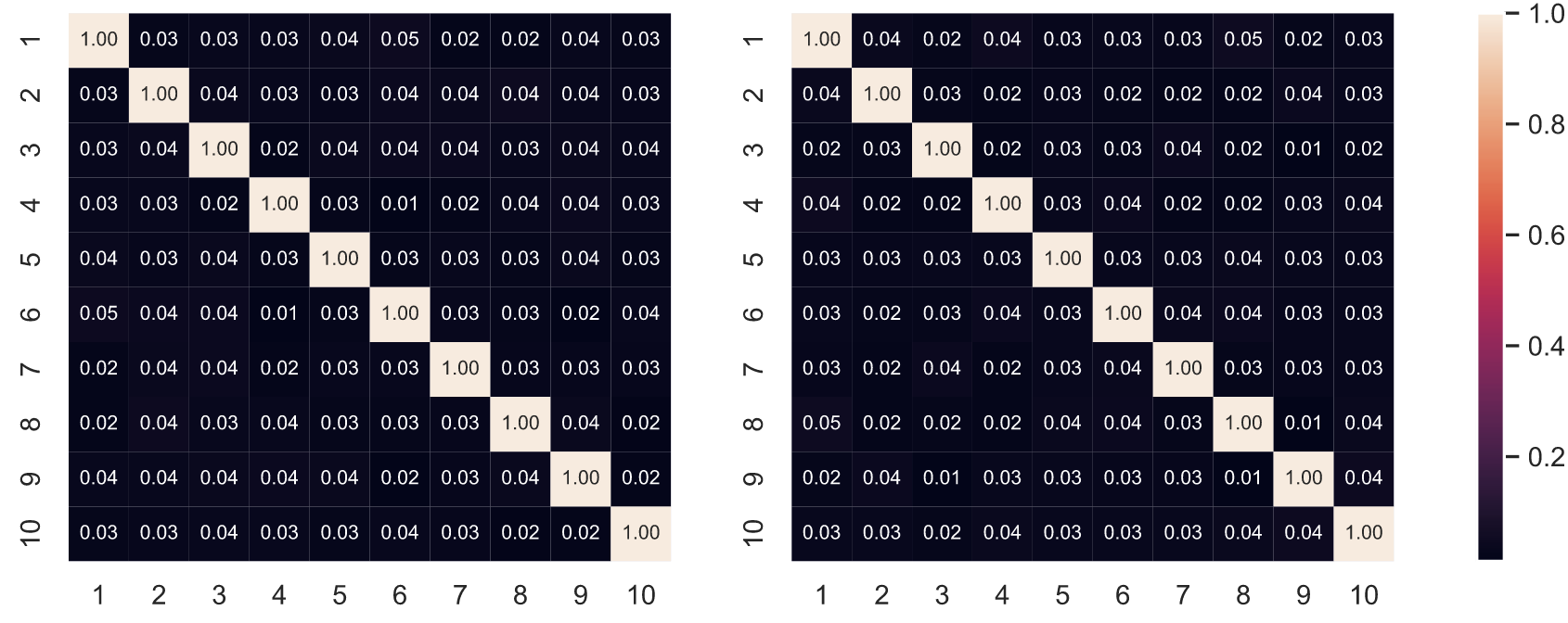}}
\vspace{-1mm}
\caption{Intersection-over-union overlap between all pairs of client masks found by {\algname} for the final ResNet18 linear layer, for $p=0.50$. Both the $s=2$ masks (left) and $s=4$ masks (right) exhibit significant diversity.}
\label{figure2}
\end{center}
\vspace{-8mm}
\end{figure}

Illustrated in Figure \ref{figure2}, the set of masks found in ResNet18's linear layer during GradLTN are significantly different from one another, indicating a high degree of personalization between clients. The resulting increased performance of {\algname} suggests that personalizing individual parameters as opposed to full layers is beneficial for FL.

\begin{figure}
\begin{center}
\centerline{\includegraphics[width=\columnwidth]{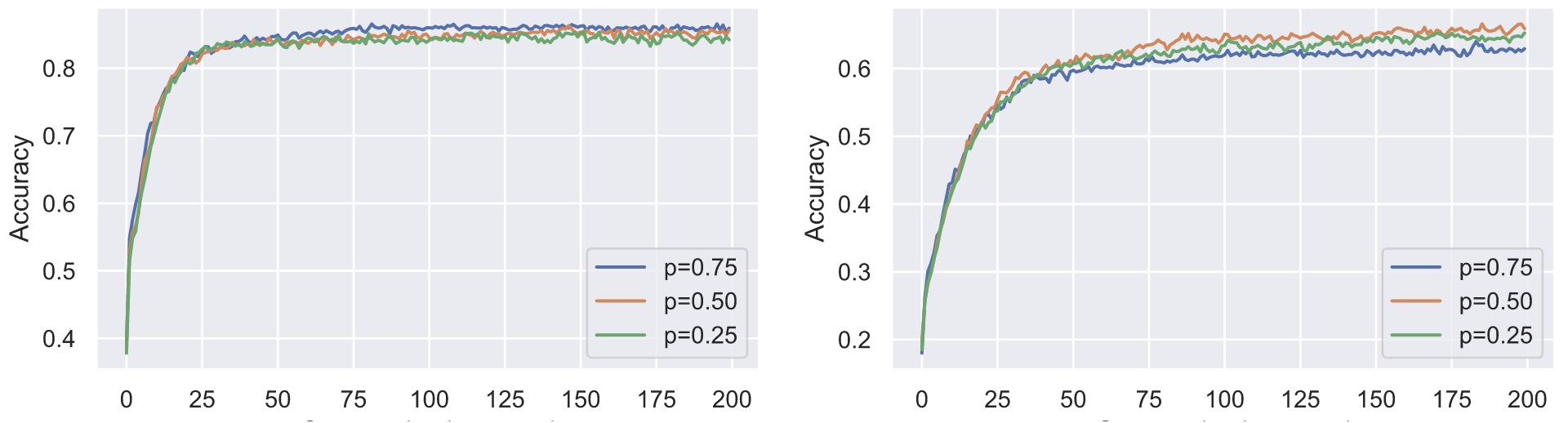}}
\vspace{-3mm}
\caption{Average test accuracies of {\algname} on non-iid client partitions of CIFAR-10 when varying the GradLTN personalization rate $p$ for $s=2$ (left) and $s=4$ (right). }
\label{figure1}
\end{center}
\vspace{-8mm}
\end{figure}

\subsection{Effect of Personalization Rate}

In Figure \ref{figure1}, we find that in the $s=2$ task, there is minimal variation in test accuracy when changing the personalization rate during {\algname}. It is possible that the process by which personalizable parameters are identified via GradLTN and LocalAlt is performed may be strong enough to perform well in the $s=2$ task despite the low data regime. However, for $s=4$, both high ($p=0.75$) and low ($p=0.25$) personalization rates perform slightly worse than $p=0.50$. Although the rate does not directly reflect the final percentage of 
 parameters frozen/personalized, $p=0.5$ may represent a middle ground for which personalizing and averaging a similar number of parameters is optimal.

\section{Conclusion \& Future Directions}
We propose {\algname}, a method for personalized federated learning that personalizes client architectures during training with the Gradient-Based Lottery Ticket Hypothesis. We demonstrate promising results on CIFAR-10, surpassing prior personalized FL and pruning-based LTH approaches in the full-participation, low-client setting. For future work, we aim to expand our method and perform extensive studies on varying the personalization rate under different settings, and apply this technique to additional datasets, such as EMNIST \cite{cohen2017emnist}, Fashion MNIST \cite{xiao2017fashion}, and CINIC10 \cite{he2020fedml}.

\section{Acknowledgements}
We would like to thank the anonymous reviewers for their valuable comments. We are thankful for the help of Chulin Xie and Wenxuan Bao for their valuable advising and support on this project. This research is part of the Delta research computing project, which is supported by the National Science Foundation (award OCI 2005572), and the State of Illinois. Delta is a joint effort of the University of Illinois at Urbana-Champaign and its National Center for Supercomputing Applications. We would also like to thank Amazon, Microsoft, and NCSA for providing conference travel funding, as well as ICML for providing registration funding. 

\end{spacing}

{
\bibliographystyle{icml2023}
\bibliography{optifine}

\begin{thebibliography}{23}
\providecommand{\natexlab}[1]{#1}
\providecommand{\url}[1]{\texttt{#1}}
\expandafter\ifx\csname urlstyle\endcsname\relax
  \providecommand{\doi}[1]{doi: #1}\else
  \providecommand{\doi}{doi: \begingroup \urlstyle{rm}\Url}\fi

\bibitem[Arivazhagan et~al.(2019)Arivazhagan, Aggarwal, Singh, and
  Choudhary]{arivazhagan2019federated}
Arivazhagan, M.~G., Aggarwal, V., Singh, A.~K., and Choudhary, S.
\newblock Federated learning with personalization layers, 2019.

\bibitem[Cohen et~al.(2017)Cohen, Afshar, Tapson, and van
  Schaik]{cohen2017emnist}
Cohen, G., Afshar, S., Tapson, J., and van Schaik, A.
\newblock Emnist: extending mnist to handwritten letters.
\newblock In \emph{2017 International Joint Conference on Neural Networks
  (IJCNN)}, 2017.

\bibitem[Collins et~al.(2021)Collins, Hassani, Mokhtari, and
  Shakkottai]{collins2021exploiting}
Collins, L., Hassani, H., Mokhtari, A., and Shakkottai, S.
\newblock Exploiting shared representations for personalized federated
  learning.
\newblock In \emph{International Conference on Machine Learning}, pp.\
  2089--2099. PMLR, 2021.

\bibitem[Deng et~al.(2009)Deng, Dong, Socher, Li, Li, and Fei-Fei]{5206848}
Deng, J., Dong, W., Socher, R., Li, L.-J., Li, K., and Fei-Fei, L.
\newblock Imagenet: A large-scale hierarchical image database.
\newblock In \emph{2009 IEEE Conference on Computer Vision and Pattern
  Recognition}, pp.\  248--255, 2009.
\newblock \doi{10.1109/CVPR.2009.5206848}.

\bibitem[Dinh et~al.(2020)Dinh, Tran, and Nguyen]{dinh2020personalized}
Dinh, T.~C., Tran, N., and Nguyen, J.
\newblock Personalized federated learning with moreau envelopes.
\newblock In \emph{Advances in Neural Information Processing Systems},
  volume~33, pp.\  21394--21405, 2020.

\bibitem[Frankle \& Carbin(2019)Frankle and Carbin]{frankle2019lottery}
Frankle, J. and Carbin, M.
\newblock The lottery ticket hypothesis: Finding sparse, trainable neural
  networks.
\newblock In \emph{International Conference on Learning Representations}, 2019.

\bibitem[He et~al.(2020)He, Li, So, Zhang, Wang, Wang, Vepakomma, Singh, Qiu,
  Shen, Zhao, Kang, Liu, Raskar, Yang, Annavaram, and Avestimehr]{he2020fedml}
He, C., Li, S., So, J., Zhang, M., Wang, H., Wang, X., Vepakomma, P., Singh,
  A., Qiu, H., Shen, L., Zhao, P., Kang, Y., Liu, Y., Raskar, R., Yang, Q.,
  Annavaram, M., and Avestimehr, S.
\newblock Fedml: A research library and benchmark for federated machine
  learning.
\newblock \emph{CoRR}, 2020.

\bibitem[He et~al.(2015)He, Zhang, Ren, and Sun]{he2015deep}
He, K., Zhang, X., Ren, S., and Sun, J.
\newblock Deep residual learning for image recognition, 2015.

\bibitem[Hsieh et~al.(2020)Hsieh, Phanishayee, Mutlu, and
  Gibbons]{hsieh2020noniid}
Hsieh, K., Phanishayee, A., Mutlu, O., and Gibbons, P.~B.
\newblock The non-iid data quagmire of decentralized machine learning, 2020.

\bibitem[Kirkpatrick et~al.(2017)Kirkpatrick, Pascanu, Rabinowitz, Veness,
  Desjardins, Rusu, Milan, Quan, Ramalho, Grabska-Barwinska, Hassabis, Clopath,
  Kumaran, and Hadsell]{Kirkpatrick_2017}
Kirkpatrick, J., Pascanu, R., Rabinowitz, N., Veness, J., Desjardins, G., Rusu,
  A.~A., Milan, K., Quan, J., Ramalho, T., Grabska-Barwinska, A., Hassabis, D.,
  Clopath, C., Kumaran, D., and Hadsell, R.
\newblock Overcoming catastrophic forgetting in neural networks.
\newblock \emph{Proceedings of the National Academy of Sciences}, 114\penalty0
  (13):\penalty0 3521--3526, mar 2017.
\newblock \doi{10.1073/pnas.1611835114}.
\newblock URL \url{https://doi.org/10.1073%2Fpnas.1611835114}.

\bibitem[Lee et~al.(2023)Lee, Chen, Tajwar, Kumar, Yao, Liang, and
  Finn]{lee2023surgical}
Lee, Y., Chen, A.~S., Tajwar, F., Kumar, A., Yao, H., Liang, P., and Finn, C.
\newblock Surgical fine-tuning improves adaptation to distribution shifts,
  2023.

\bibitem[Li et~al.(2020)Li, Sun, Wang, Duan, Li, Chen, and Li]{li2020lotteryfl}
Li, A., Sun, J., Wang, B., Duan, L., Li, S., Chen, Y., and Li, H.
\newblock Lotteryfl: Personalized and communication-efficient federated
  learning with lottery ticket hypothesis on non-iid datasets, 2020.

\bibitem[Li et~al.(2021{\natexlab{a}})Li, Hu, Beirami, and Smith]{li2021ditto}
Li, T., Hu, S., Beirami, A., and Smith, V.
\newblock Ditto: Fair and robust federated learning through personalization.
\newblock In \emph{International Conference on Machine Learning}, pp.\
  6357--6368. PMLR, 2021{\natexlab{a}}.

\bibitem[Li et~al.(2021{\natexlab{b}})Li, JIANG, Zhang, Kamp, and Dou]{lifedbn}
Li, X., JIANG, M., Zhang, X., Kamp, M., and Dou, Q.
\newblock Fedbn: Federated learning on non-iid features via local batch
  normalization.
\newblock In \emph{International Conference on Learning Representations},
  2021{\natexlab{b}}.

\bibitem[Liang et~al.(2020)Liang, Liu, Ziyin, Allen, Auerbach, Brent,
  Salakhutdinov, and Morency]{liang2020think}
Liang, P.~P., Liu, T., Ziyin, L., Allen, N.~B., Auerbach, R.~P., Brent, D.,
  Salakhutdinov, R., and Morency, L.-P.
\newblock Think locally, act globally: Federated learning with local and global
  representations.
\newblock \emph{arXiv preprint arXiv:2001.01523}, 2020.

\bibitem[Liu et~al.(2022)Liu, Hu, Wu, and Smith]{liu2022privacy}
Liu, Z., Hu, S., Wu, Z.~S., and Smith, V.
\newblock On privacy and personalization in cross-silo federated learning,
  2022.

\bibitem[McCloskey \& Cohen(1989)McCloskey and
  Cohen]{McCloskey1989CatastrophicII}
McCloskey, M. and Cohen, N.~J.
\newblock Catastrophic interference in connectionist networks: The sequential
  learning problem.
\newblock \emph{Psychology of Learning and Motivation}, 24:\penalty0 109--165,
  1989.

\bibitem[McMahan et~al.(2017)McMahan, Moore, Ramage, Hampson, and
  y~Arcas]{mcmahan2017}
McMahan, B., Moore, E., Ramage, D., Hampson, S., and y~Arcas, B.~A.
\newblock Communication-efficient learning of deep networks from decentralized
  data.
\newblock In \emph{Proc. of Int’l Conf. Artificial Intelligence and
  Statistics (AISTATS)}, Apr 2017.

\bibitem[Mugunthan et~al.(2022)Mugunthan, Lin, Gokul, Lau, Kagal, and
  Pieper]{10.1007/978-3-031-19775-8_5}
Mugunthan, V., Lin, E., Gokul, V., Lau, C., Kagal, L., and Pieper, S.
\newblock Fedltn: Federated learning for sparse and personalized lottery
  ticket networks.
\newblock In Avidan, S., Brostow, G., Ciss{\'e}, M., Farinella, G.~M., and
  Hassner, T. (eds.), \emph{Computer Vision -- ECCV 2022}, pp.\  69--85, Cham,
  2022. Springer Nature Switzerland.
\newblock ISBN 978-3-031-19775-8.

\bibitem[Oh et~al.(2022)Oh, Kim, and Yun]{oh2022fedbabu}
Oh, J., Kim, S., and Yun, S.-Y.
\newblock Fed{BABU}: Toward enhanced representation for federated image
  classification.
\newblock In \emph{International Conference on Learning Representations}, 2022.
\newblock URL \url{https://openreview.net/forum?id=HuaYQfggn5u}.

\bibitem[Pillutla et~al.(2022)Pillutla, Malik, Mohamed, Rabbat, Sanjabi, and
  Xiao]{pillutla2022federated}
Pillutla, K., Malik, K., Mohamed, A., Rabbat, M., Sanjabi, M., and Xiao, L.
\newblock Federated learning with partial model personalization.
\newblock In \emph{International Conference on Machine Learning}, 2022.

\bibitem[Sheller et~al.(2020)Sheller, Edwards, Reina, Martin, and
  Bakas]{sheller2020federated}
Sheller, M.~J., Edwards, B., Reina, G.~A., Martin, J., and Bakas, S.
\newblock Federated learning in medicine: facilitating multi-institutional
  collaborations without sharing patient data.
\newblock \emph{Scientific Reports}, 10\penalty0 (12598), 2020.
\newblock \doi{10.1038/s41598-020-69250-1}.
\newblock URL \url{https://doi.org/10.1038/s41598-020-69250-1}.

\bibitem[Xiao et~al.(2017)Xiao, Rasul, and Vollgraf]{xiao2017fashion}
Xiao, H., Rasul, K., and Vollgraf, R.
\newblock Fashion-mnist: a novel image dataset for benchmarking machine
  learning algorithms.
\newblock \emph{CoRR}, 2017.

\end{thebibliography}
}

\end{document}